\documentclass[runningheads]{llncs}
\usepackage{graphicx}
\usepackage{makeidx}  % allows for indexgeneration

\usepackage{amsmath,graphicx}
\usepackage{amssymb}
\usepackage{tabulary}
\usepackage{tabularx}
\usepackage{floatflt}
\usepackage{float}
\usepackage{todonotes}
\usepackage{color}

\usepackage{amsmath}
\usepackage{amssymb}

\usepackage{multirow} 
\usepackage{floatflt}
\usepackage{url}
\usepackage{epstopdf}
\usepackage{makecell}
\usepackage{xcolor}
\definecolor{darkgreen}{RGB}{0, 190, 0}

\newcommand{\bfx}{\mathbf{x}}
\newcommand{\mdiag}{\text{{diag}}}

\newcommand{\bsigma}{\mathbf{b}}
 
\newcommand{\argmin}{\operatornamewithlimits{argmin}} 
\newcommand{\matr}[1]{\mathbf{#1}}
\newcommand{\vectr}[1]{\mathbf{#1}}
\newcommand*{\tran}{^{\mkern-1.5mu\mathsf{T}}}

\newcommand{\vvv}[1]{{{#1}}} 
\newfloat{myalg}{thp}{lop}
\floatname{myalg}{\textbf{Algorithm}}

\begin{document}
	\title{Image Reconstruction via Variational Network\\ for Real-Time Hand-Held Sound-Speed Imaging}
	\titlerunning{Real-Time SoS Reconstruction with Variational Network}
	\author{Valery Vishnevskiy\and Sergio J Sanabria\and Orcun Goksel}  
	\authorrunning{V. Vishnevskiy et al.}
    \institute{Computer-assisted Applications in Medicine Group, 
		ETH Zurich, Switzerland
	}
	\maketitle              % typeset the header of the contribution

	\begin{abstract}
Speed-of-sound is a biomechanical property for quantitative tissue differentiation, with great potential as a new ultrasound-based image modality. A conventional ultrasound array transducer can be used together with an acoustic mirror, or so-called reflector, to reconstruct sound-speed images from time-of-flight measurements to the reflector collected between transducer element pairs, which constitutes a challenging problem of limited-angle computed tomography.
For this problem, we herein present a variational network based image reconstruction architecture that is based on optimization loop unrolling, and provide an efficient training protocol of this network architecture on fully synthetic inclusion data.
Our results indicate that the learned model presents good generalization ability, being able to reconstruct images with significantly different statistics compared to the training set.
Complex inclusion geometries were shown to be successfully reconstructed, also improving over the prior-art by 23\% in reconstruction error and by 10\% in contrast on synthetic data.
In a phantom study, we demonstrated the detection of multiple inclusions that were not distinguishable by prior-art reconstruction, meanwhile improving the contrast by 27\% for a stiff inclusion and by 219\% for a soft inclusion.
Our reconstruction algorithm takes approximately 10\,ms, enabling its use as a real-time imaging method on an ultrasound machine, for which we are demonstrating an example preliminary setup herein.

	\end{abstract}
	\section{Introduction}
	Speed-of-sound (SoS) ultrasound computed tomography (USCT) is a promising image modality, which generates maps of speed of sound in tissue as an imaging biomarker. Potential clinical applications are differentiation of breast tumorous lesions \cite{Duric10}, breast density assessment \cite{Sanabria2018,Sak15}, staging of musculoskeletal \cite{Qu2017} and non-alcoholic fatty liver disease \cite{Imbault2018}, amongst others.
	For this, a set of time of flight (ToF) measurements through the tissue between pairs of transmit/receive elements of an ultrasonic array can be used for a tomographic reconstruction. Various 2D ad 3D acquisition setups have been proposed, including circular or dome-shaped transducer geometries, which provide multilateral set of measurements that are convenient for reconstruction methods~\cite{jirik2012sound} but costly to manufacture and cumbersome in use.
	Hand-held reflector based setup~\cite{krueger1998limited,sanabria2016hand} depicted in Fig.\,\ref{fig:geo_data}a uses a conventional portable ultrasound probe to measure ToF via wave reflections of a plate placed on the opposite side of the sample.
	Despite its simplicity, such a setup results in limited-angle (LA) CT, which requires prior assumptions and suitable regularization and numerical optimization techniques to produce meaningful reconstructions~\cite{sanabria2016hand}.
	Such optimization techniques may not be guaranteed to converge, are often slow in runtime, and involve parameters that are difficult to set.
	
	In this paper, we propose a problem-specific variational network~\cite{hammernik2018learning,adler2017solving} for limited-angle SoS reconstruction, with parameters learned from numerous forward simulations.
	Contrary to machine learning methods based on sinogram inpainting~\cite{tovey2018directional} and reconstruction artefact removal~\cite{hammernik2017deep} for LA-CT, we learn reconstruction process end-to-end, and show that it allows to qualitatively improve conventional reconstruction.
	
	\section{Methods}
\begin{figure}[t]
    \scriptsize
		\begin{minipage}[t]{0.48\linewidth}
			\centering
			%\vfill
			\centerline{\includegraphics[height=4cm]{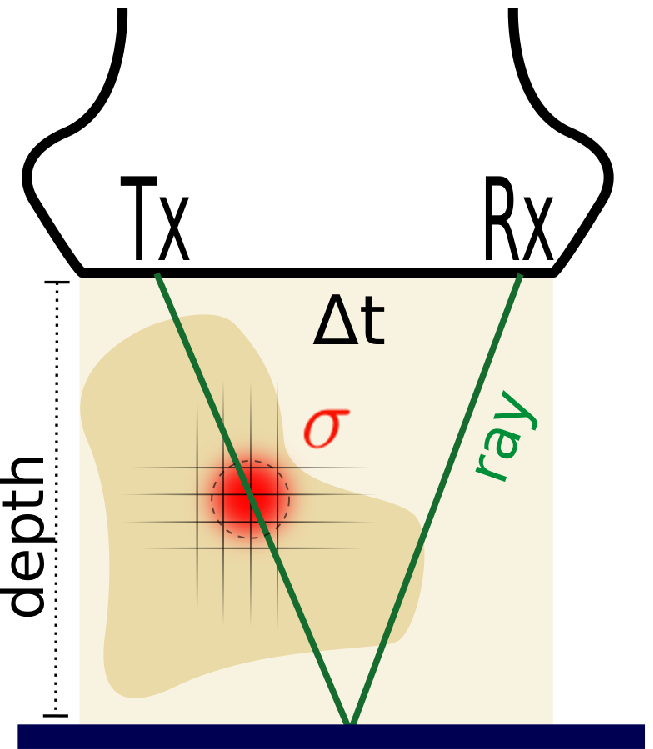}}
			%  \vspace{1.5cm}
			\centerline{(a)}\medskip
		\end{minipage}
		\hspace{4mm}
        \begin{minipage}[t]{0.48\linewidth}
        \includegraphics[height=4cm]{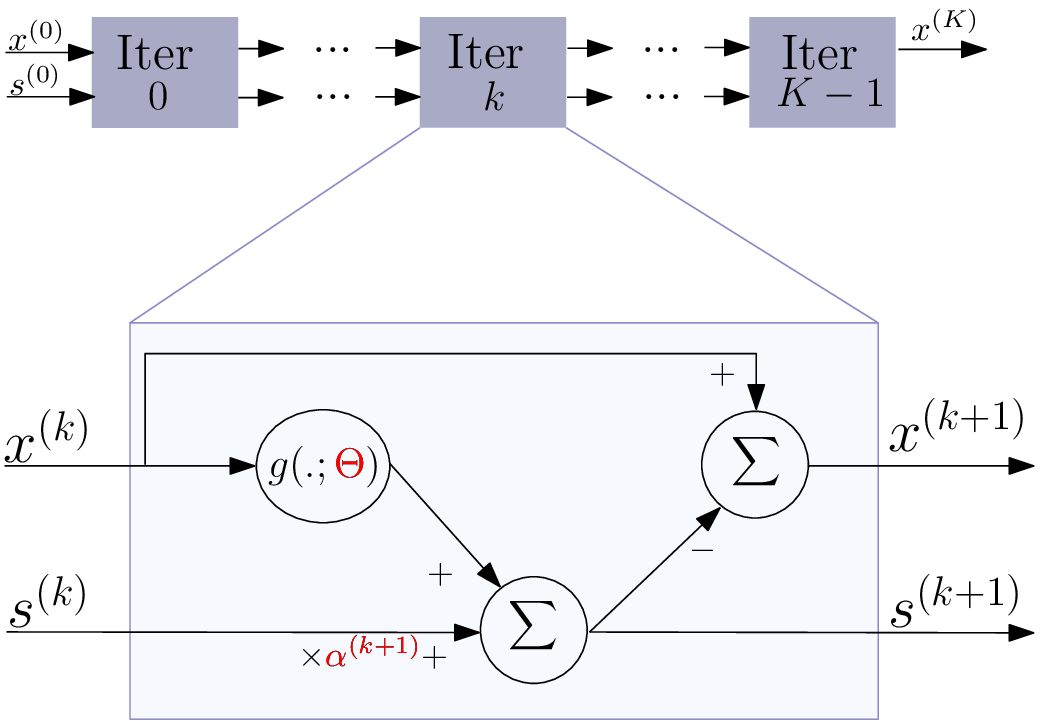}
        \centerline{(b)}\medskip
        \end{minipage}
        \\

        \begin{minipage}[t]{0.9\linewidth}
			\centering
			%\vfill
			\hspace{60pt} Synthetic inclusions $(\mathcal{T}\, )$
            \hspace{60pt} Geometric shapes $(\mathcal{P})$\\
			\hspace{5mm}\includegraphics[height=3.5cm]{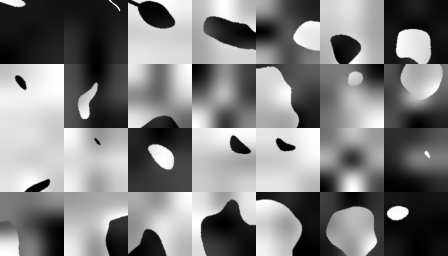}
			\hspace{10mm}
			\includegraphics[height=3.5cm]{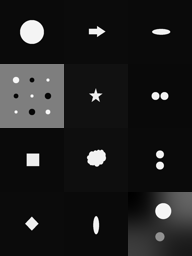}
			%  \vspa8ce{1.5cm}
			\centerline{(c)}\medskip
		\end{minipage}
		\vspace{-4mm}
        
		\caption{(a)~Acquisition setup and ray tracing discretization. 
			\;
            (b)~\vvv{Structure of variational network; tunable parameters of the layer are highlighted in red.}
            \;
			(c)~Samples from synthetic training set $\mathcal{T}$ and testing set of geometric primitives $\mathcal{P}$. }
		\label{fig:geo_data}
	\end{figure}

	Using the wave reflection tracking algorithm described in~\cite{sanabria2016hand}, we measure the 
	ToF $\Delta t$ between transmit (Tx) and receive (Rx) transducers in a $M$$=$$128$ element linear ultrasound array (see Fig.\,\ref{fig:geo_data}a).
	Discretizing corresponding ray paths using a Gaussian sampling kernel, the inverse of ToF can be expressed as a linear combination of tissue slowness values $x$ [s/m], i.e.\ 
	$(\Delta t)^{-1}=\sum_{i\in\text{Ray}}l_i x_i$.
	Considering a Cartesian $n_1$$\times$$n_2$=$P$ grid, we define the forward model
	\begin{equation}
		\label{eq:fwd_prob}
		\bsigma = \mdiag(\vectr{m})\matr{L}\bfx + \mathcal{N}(\vectr{0}, \sigma_N\matr{I}),
	\end{equation}
	where $\vectr{x}\in\mathbb{R}^{P}$ is the inverse SoS (slowness) map,
	$\matr{L}\in\mathbb{R}^{ M^2\times P}$ is a sparse path matrix defined by acquisition geometry and discretization scheme,
	$\vectr{m}\in\{0,1\}^{M^2}$ is the undersampling mask with zeros indicating a missing (e.g., unreliable) ToF measurement between a corresponding Tx-Rx pair, 
	and $\vectr{b}\in\mathbb{R}^{M^2}$ is a zero-filled 
	%\todo{why is b zero padded} 
	vector of measured inverse ToFs $(\Delta t)^{-1}$.
	Reconstructing a slowness map $\bfx$ is a process inverse to~(\ref{eq:fwd_prob}) and can be posed as the following convex optimization problem:
	\begin{equation}
		\label{eq: var_rec}
		\hat{\bfx}(\bsigma, \vectr{m}; \lambda, \boldsymbol{\nabla}) = \argmin_{\bfx}\, \|\mdiag(\vectr{m})\matr{L}\bfx - \bsigma\|_1 + \lambda\|\boldsymbol{\nabla}\bfx\|_1,
	\end{equation}
	which we solve using ADMM~\cite{boyd2011distributed} algorithm with Cholesky factorization.  
	Here $\boldsymbol{\nabla}$~is a matrix, and $\lambda$ is the regularization weight. 
	
	It is common to choose regularization matrix $\boldsymbol{\nabla}_\text{TV}$ that implements spatial gradients on Cartesian grid, yielding the total variation (TV) regularization~\cite{Rudin1992259}, which allows to efficiently recover sharp image boundaries, but can introduce signal underestimation and staircase artefacts that are amplified by the limited-angle acquisition.
	In attempt to remedy this problem, one can delicately construct a set of image filters that will penalize problem-specific reconstruction artefacts.
	We follow~\cite{sanabria2016hand} and use regularization matrix $\boldsymbol{\nabla}_\text{MATV}$ that implements convolution with the set of weighted directional gradient operators. This weights regularization according to known wave path information, such that the locations with information from a narrower angular range are regularized more.
	
	\subsection{Variational Network}
	Variational networks (VN) is a class of deep learning methods that incorporate a parametrized prototype of a reconstruction algorithm in differentiable manner. 
	A successful VN architecture proposed by Hammernik et al.\ in~\cite{hammernik2018learning} for undersampled MRI reconstruction unrolls a fixed number of iterations of the gradient descent (GD) algorithm applied to a virtual optimization-based reconstruction problem. 
	By unrolling the iterations of the algorithm into network layers \vvv{(see Fig.~\ref{fig:geo_data}b)}, the output is expressed as a formula parametrized by the regularization parameters and step lengths of this GD algorithm. 
	The parameters are then tuned on retrospectively undersampled training data.
	
	In contrast to discrete Fourier transform, the design matrix of LA-CT is poorly conditioned, which compromises the performance of conventional GD.
	Therefore, we propose to enhance the VN in the following ways: (i) unroll GD with momentum, (ii) add left diagonal preconditioner $\vectr{p}^{(k)}\in\mathbb{R}^{M^2}$ for the path matrix $\matr{L}$, (iii) use adaptive data consistency term $\varphi_\text{d}^{(k)}$, and (iv) allow spatial filter weighting $\matr{w}_i^{(k)}\in\mathbb{R}^{P}$.
	The resulting reconstruction network is defined in Algorithm~\ref{alg:varnet} with tunable parameters $\Theta$, where each of $K$ variational layers contains $N_f$ convolution matrices $\matr{D}=\matr{D}(\vectr{d})$ with $N_c\times N_c$ kernels $\vectr{d}$ that are ensured to be zero-centered unit-norm via re-parametrization: 
	$\vectr{d}=(\vectr{d}'-\langle\vectr{d}'\rangle )/\|(\vectr{d}'-\langle\vectr{d}'\rangle )\|_2$, where $\langle.\rangle$ denotes mean value of the vector.
	Each filter $\matr{D}$ is associated with its potential function $\varphi_{\text{r}}\{.\}$ that is parametrized via cubic interpolation of control knots ${\boldsymbol\phi}_\text{r}\in\mathbb{R}^{N_g}$ placed on Cartesian grid on $[-r, r]$ interval. 
	Data term potentials ${\varphi}_\text{d}\{.\}$ are defined in the same way.
	The network is trained to minimize $\ell_1$-norm of the reconstruction error on the training set $\mathcal{T}$:
	\begin{equation}
		\min_\Theta \mathop{\mathbb{E}}_{ \{\bsigma, \vectr{m}, \bfx^\star\} \in \mathcal{T}} 
		\|\mathcal{V}(\bsigma, \vectr{m}; \,\Theta) - \bfx^\star\|_1.
	\end{equation}

	\textbf{Training} dataset $\mathcal{T}$ is generated using fixed acquisition geometry with reflector depth equal to transducer array width. 
	High-resolution (HR) 256$\times$256 synthetic inclusion masks are produced by applying smooth deformation to an ellipse with random center, eccentricity, and radius. 
	Two smooth slowness maps with random values from $[1/1650, 1/1350]$ interval are then blended with this inclusion mask, yielding a final slowness map $\bfx^\star_\text{HR}$ (see Fig.\,\ref{fig:geo_data}c). 
	The chosen range corresponds to observed SoS values for breast tissues of different densities and tumorous inclusions of different pathologies~\cite{duric2007detection}.
	%Random incoherent undersmapling mask $\vectr{m}$ is generated.
	Forward path matrix $\matr{L}_\text{HR}$ and random incoherent undersmapling mask $\vectr{m}$ are used to generate noisy inverse ToF vector $\vectr{b}$ according to model~(\ref{eq:fwd_prob}) with $\sigma_N$=$2\cdot10^{-8}$.
	Finally, we downsample $\vectr{x}^\star_\text{HR}$ to $n_1$$\times$$n_2$ size yielding the ground truth map $\vectr{x}^\star$.
	About 10\% of maps did not contain inclusions.
	For each reconstruction problem the path matrix $\matr{L}$ is normalized with its largest singular value,
	and inverse ToF are centered and scaled: $\bsigma' = \bsigma - (\langle\bsigma\rangle /\langle  \matr{L}\vectr{1}\rangle)\matr{L}\vectr{1}$, 
	$\tilde{\bsigma} = \bsigma' / \text{std}(\bsigma')$.
	
	The configuration of networks were the following: $K$=10, $N_f$=50, $N_c$=5, $n_1$=$n_2$=64, $N_g$=55. 
	All parameters were initialized from $\mathcal{U}(0, 1)$.
	We refer to this architecture as VNv4. Ablating spatial filter weighting $\vectr{w}_i^{(k)}$ from VNv4, we get VNv3; additionally ablating adaptive data potentials $\varphi_\text{d}^{(k)}$, we get VNv2; further ablating preconditioner $\vectr{p}^{(k)}$, we get VNv1; and eventually unrolling GD without momentum, VNv0.
	For tuning the aforementioned models we used $10^5$ iterations of Adam algorithm~\cite{kingma2014adam} with  learning rate $10^{-3}$ and batch size 25.
    \vvv{Every 5000 iterations we readjust potential function's interval range $r$ by setting it to the maximal observed value of the corresponding activation function argument.}
	
	\begin{myalg}[t]
		%footnotesize
        %\scriptsize
		%\hspace{-5mm}
		\begin{tabular}{m{\linewidth}}
			\hline \\
			\vspace{-0.55cm}
			{Input: $\bsigma$~--- inverse ToF,\, $\vectr{m}$~--- undersampling mask }\\
            \vspace{-0.45cm}
			{Parameters: $\Theta = \{\boldsymbol{\phi}^{(k)}_\text{d},\boldsymbol{\phi}^{(k)}_{\text{r},i}, \vectr{p}^{(k)}, \vectr{w}_i^{(k)}, \matr{D}_i^{(k)}, \alpha^{(k)}\}_{i=1,\dots,N_f,\; k=1,\dots,K}$}\\
            \vspace{-0.35cm}
			{$\bfx^{(0)} \leftarrow \alpha^{(0)}\matr{L}\tran\bsigma,\quad\vectr{s}^{(0)}\leftarrow\vectr{0}$}\\
            \vspace{-0.35cm}
			{\texttt{\textbf{for}} $k:=0$ to $K-1$}\\
            \vspace{-0.35cm}
			{\quad $\vectr{g}^{(k)}\leftarrow \matr{L}\tran \mdiag(\vectr{p}^{(k)})\mdiag(\vectr{m})\, \varphi_\text{d}^{(k)}\left\{\mdiag(\vectr{m})\mdiag(\vectr{p}^{(k)}) \left(\matr{L}\bfx^{(k)} - \bsigma\right) \right\} + $ }\\
			{\quad\quad\hspace{40pt}\quad $			\sum_{i=1,\dots,N_f}\left(\matr{D}^{(k)}_{i}\right)\tran\mdiag(\vectr{w}^{(k)}_i)\,\varphi^{(k)}_{\text{r},i} \left\{ \mdiag(\vectr{w}^{(k)}_i) \matr{D}^{(k)}_{i}\bfx^{(k)}\right\}$}\\
			{\quad $\vectr{s}^{(k+1)}\leftarrow \alpha^{(k+1)}\vectr{s}^{(k)} + \vectr{g}^{(k)}$}\\
			{\quad $\vectr{x}^{(k+1)}\leftarrow \vectr{x}^{(k)} - \vectr{s}^{(k+1)}$}\\
			{Output: reconstructed image $\mathcal{V}(\bsigma, \vectr{m}; \,\Theta) := \bfx^{(K)}$}\\
			\hline
		\end{tabular}
		\vspace{-2mm}
		\caption{
			Proposed variational reconstruction network model VNv4.
		}
        %\vspace{-5mm}
		\label{alg:varnet}
	\end{myalg}

	\section{Results}
\begin{figure}[H]
	{\tiny Ground Truth\hspace{18pt} TV\hspace{19pt} MA-TV
		\hspace{15pt}\fbox{{\bf VNv4}} \hspace{19pt}VNv3\hspace{22pt}VNv2\hspace{22pt}VNv1\hspace{24pt}VNv0 }\\
	\includegraphics[width=\textwidth]{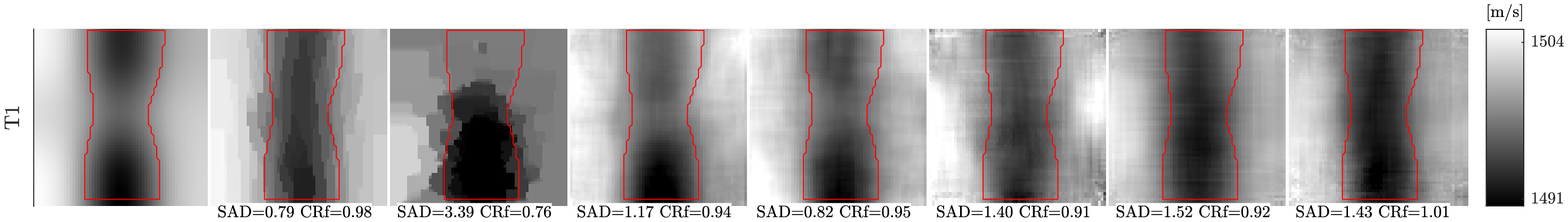}\\
	\includegraphics[width=\textwidth]{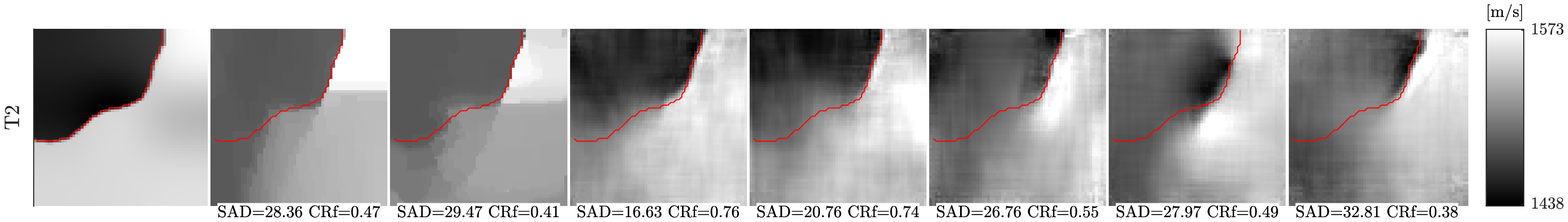}
	\includegraphics[width=\textwidth]{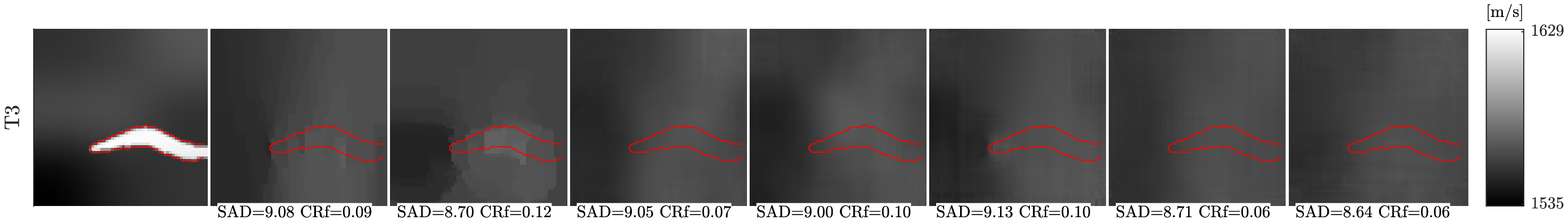}
	\includegraphics[width=\textwidth]{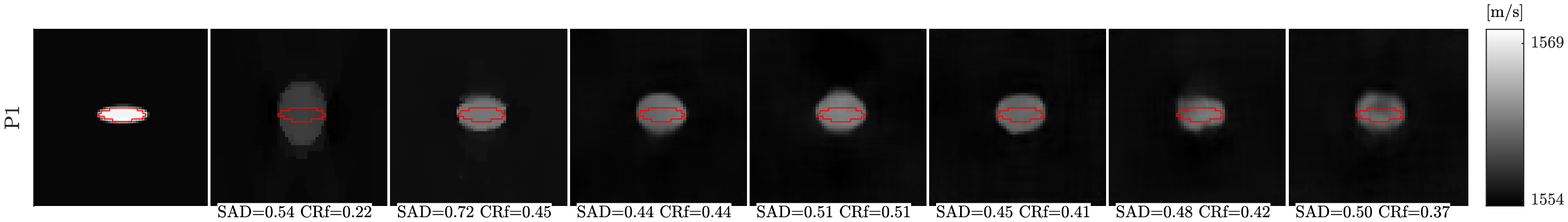}
	\includegraphics[width=\textwidth]{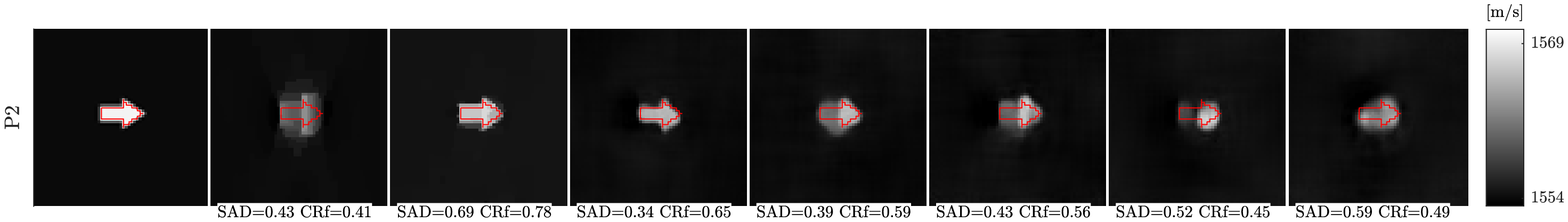}
	\includegraphics[width=\textwidth]{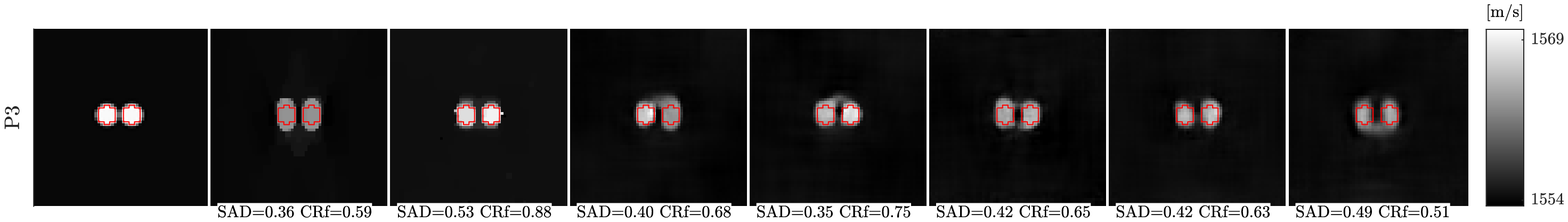}
	\includegraphics[width=\textwidth]{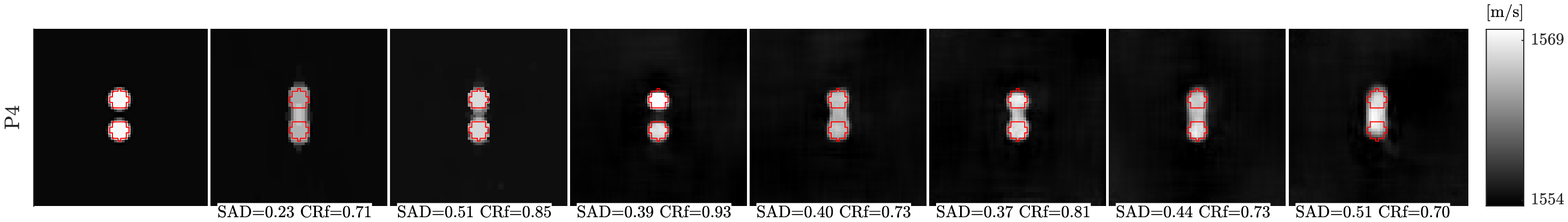}
	\includegraphics[width=\textwidth]{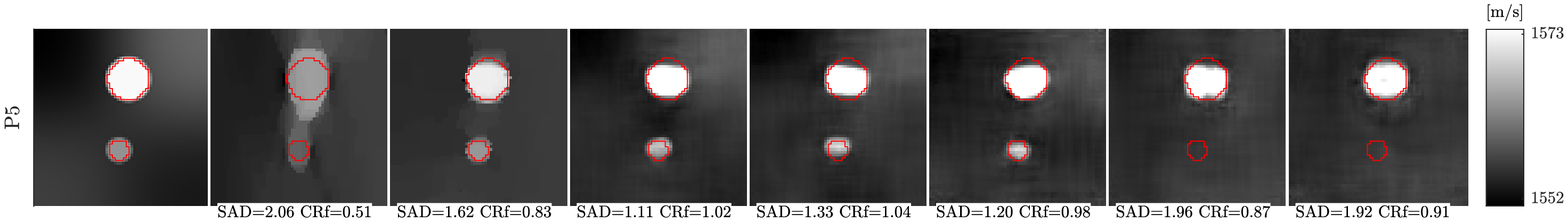}
	\caption{Sound speed reconstructions of synthetic data from sets $\mathcal{T}$, $\mathcal{P}$ and single natural image N1. 
		Inclusions are delineated with red curves.
		For each image transducer array is placed on top and reflector on bottom. 
	} 
	\label{fig:bigimg}
\end{figure}

	We compare TV and MA-TV against VN architectures on (i) 200 samples from $\mathcal{T}$ that were set aside and unseen during training, and (ii) a set $\mathcal{P}$ of 14 geometric primitives depicted in Fig.\,\ref{fig:geo_data}c, using following metrics:
	\begin{equation}
		\text{\small SAD}(\vectr{x}, \vectr{y})=\frac{\|\vectr{x}- \vectr{y}\|_1}{P}, \quad
		\text{\small CR} = \frac{2\left| \mu_\text{inc} - \mu_\text{bg} \right|}{\left|\mu_\text{inc}\right| + \left|\mu_\text{bg}\right|},\quad \text{CRf}=\frac{\text{\small  estimated CR}}{\text{\small ground truth CR}},
	\end{equation}
	where $\mu_\text{inc}$ and $\mu_\text{bg}$ are mean values in the inclusion and background regions accordingly. The optimal regularization weight $\lambda$ for TV and MA-TV algorithms was tuned to give the best (lowest) SAD on the P3 image (see Fig.\,\ref{fig:bigimg}).
	Similarly to training generation, the forward model for validation and test sets was computed on high resolution images with normal noise and 30\% undersmapling.
	
	Quantitative evaluation on synthetic data is reported in Table~\ref{tbl:res_table} and shows that the proposed VNv4 network outperforms conventional TV and MA-TV reconstruction methods both in terms of accuracy and contrast.
	Comparing VNv options, it can be seen that richer architectures performed better. 
	Fig.~\ref{fig:bigimg} shows qualitative evaluation of reconstruction methods. 
	VNv4 is able to reconstruct multiple inclusions (P5), handle smooth SoS variation (T1), and generally maintain inclusion position and geometry without hallucinating nonexistent inclusions.
    \vvv{Although for some geometries (e.g. P4) TV reconstruction has lower SAD value, VNv4 provides better contrast, which allows to separate the two inclusions.}
	As expected from the limited-angle nature of the data, highly elongated inclusions that are parallel to the reflector \vvv{either undergo axial geometric distortion~(P1)}, or could not be adequately reconstructed \vvv{(T3)} by any presented method.
   	
\subsubsection{Breast Phantom Experiment.}
\begin{figure}[H]
		{\scriptsize~\hspace{20pt} B-mode\hspace{54pt} VNv4\hspace{58pt} MA-TV
			\hspace{61pt}TV }\\
                %\centering
		\includegraphics[width=\textwidth]{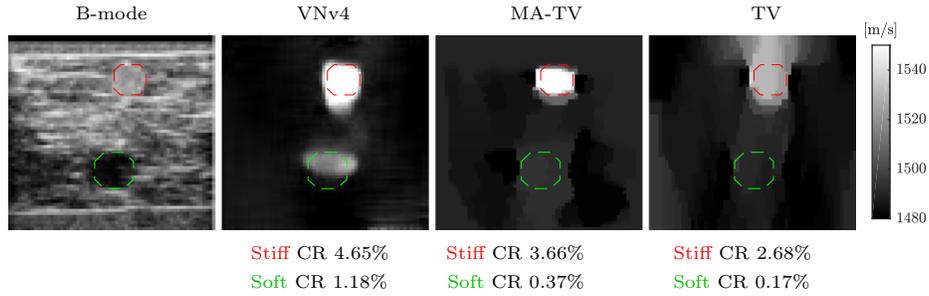}\\
		{\scriptsize \hspace*{89pt} \textcolor{red}{Stiff} CR 4.65\%\hspace{19pt} \textcolor{red}{Stiff} CR 3.66\%\hspace{31pt} \textcolor{red}{Stiff} CR 2.68\%}\\
		{\scriptsize \hspace*{89pt} \textcolor{darkgreen}{Soft} CR 1.18\%\hspace{20pt} \textcolor{darkgreen}{Soft} CR 0.37\%\hspace{31pt} \textcolor{darkgreen}{Soft} CR 0.17\%}
        \vspace{-2mm}
		\caption{Hand-held SoS mammography of the breast phantom. Stiff (red) and soft (green) inclusions were delineated in the B-mode image. 
			%\todo[inline]{Cool! I typically used hot color map for the SoS images, maybe it looks even better.}  
		} 
		\label{fig:exviv}
        \vspace{-5mm}
	\end{figure}

\begin{figure}[H]	
            \centering
			\includegraphics[width=0.82\textwidth]{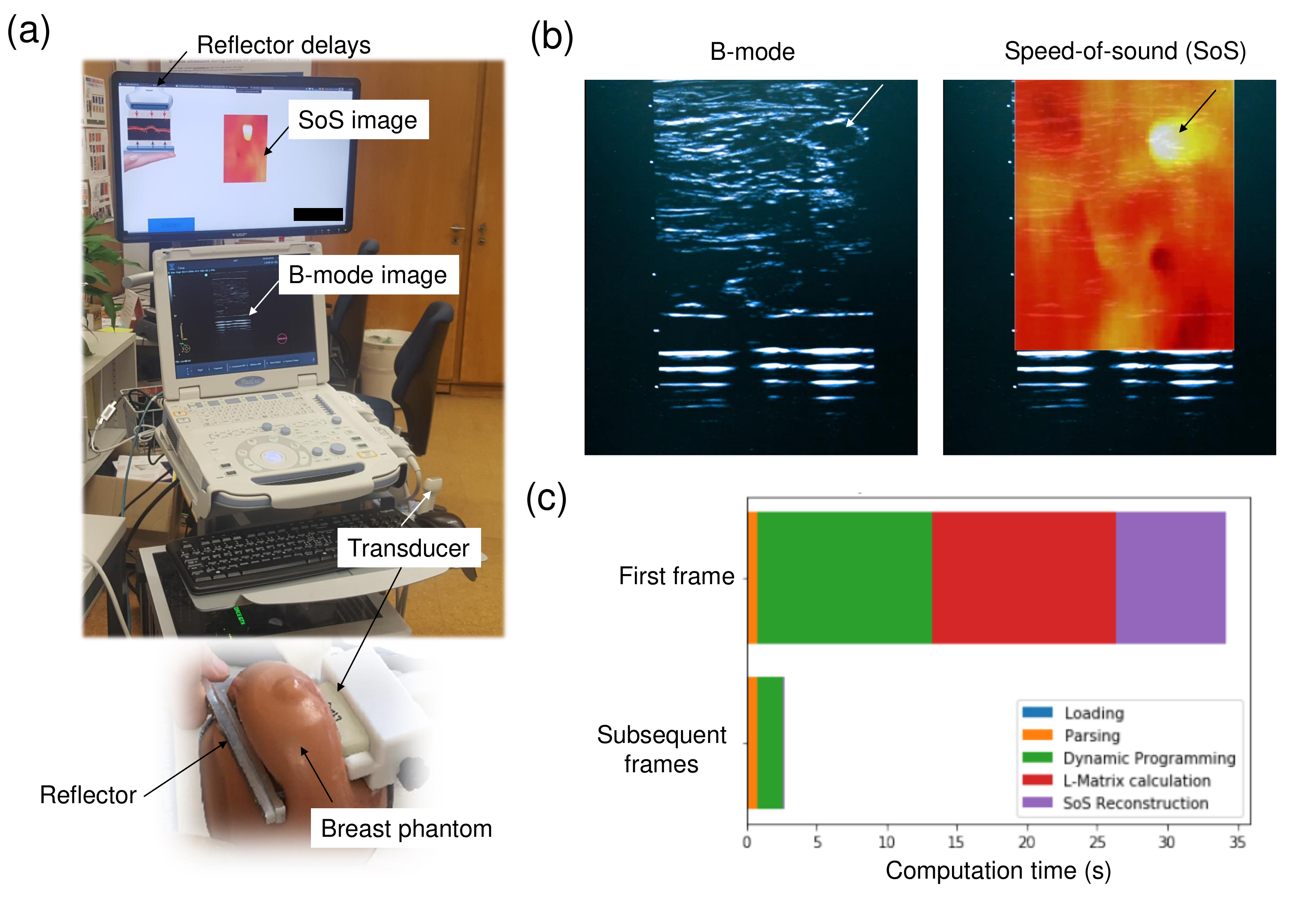}\\
            \vspace{-3mm}
			\caption{
            Live SoS imaging demonstration. 
            (a) Experimental setup. (b) Sample outputs of B-mode and SoS video feedback. A non-echogenic stiff lesion is clearly delineated in the SoS image.  (c) Computational benchmarks, also showing initialization and memory allocation times. After initialization, SoS reconstruction time is negligible compared to data transfer and reflector ToF measurement via dynamic programming \cite{sanabria2016hand}.
			}
            \label{fig:livedemo}
		\end{figure}

	We also compared the reconstruction methods using a realistic breast elastography phantom (Model 059, CIRS Inc.) that mimics glandular tissue with two lesions of different density.
	%A laptop ultrasound system 
    \vvv{Portable ultrasound} system (UF-760AG, Fukuda Denshi Inc., Tokyo, Japan) streams full-matrix RF ultrasound data over a high bandwidth link to a dedicated PC, which is used to perform USCT reconstruction and output a live SoS video feedback (cf.\ Fig.\,\ref{fig:livedemo}). 
	We used an ultrasound probe\,(FUT-LA385-12P) with 128 piezoelectric transducer elements. 
	For each frame a total of 128$\times$\!128 RF lines are generated for all Tx/Rx combinations,  at an imaging center frequency of 5\,MHz digitized at 40.96\,MHz.
	As seen in Fig.\,\ref{fig:exviv}, VNv4 qualitatively outperforms both TV and MA-TV methods, showing clearly distinguishable hard and soft lesions. 
	Run-time of MA-TV and TV algorithms on CPU is $\sim$30\,s per image, while VN reconstruction takes $\sim$0.4\,s on CPU and $\sim$0.01\,s on GPU.
	
		\section{Discussion}
	In this paper we have proposed a deep variational reconstruction network for hand-held US sound-speed imaging.
	The method is able to reconstruct various inclusion geometries both in synthetic and phantom experiments.
    \vvv{VN demonstrated good generalization ability, which suggests that unrolling even more sophisticated numerical schemes may be possible.}
	Improvements over conventional reconstruction algorithms are both qualitative and quantitative.
	The ability of method to distinguish hard and soft inclusions has great diagnostic potential in characterizing lesions in real-time.

    \begin{table}[t]%[b]
		\begin{center}
			\caption{
				SoS reconstruction measures computed on 200 validation images from training distribution $\mathcal{T}$, 14 test images from the set of geometric primitives $\mathcal{P}$.}
                \vspace{-5pt}
			\label{tbl:res_table}
			\scriptsize
			\begin{tabulary}{\linewidth}{l|C|C|C|C|C|C|C|C|C|C|C|C|C|C}
				\hline 
				\textbf{\vvv{Shape set}} & \multicolumn{2}{c|}{\textbf{TV}}& \multicolumn{2}{c|}{\textbf{MA-TV}} & \multicolumn{2}{c|}{\textbf{VNv4}} & \multicolumn{2}{c|}{\textbf{VNv3}} & \multicolumn{2}{c|}{\textbf{VNv2}} & \multicolumn{2}{c|}{\textbf{VNv1}} &
				\multicolumn{2}{c}{\textbf{VNv0}}\\
				~ & SAD & CRf& SAD & CRf& SAD & CRf& SAD & CRf& SAD & CRf& SAD & CRf& SAD & CRf \\
				\hline\hline
				\makecell[l]{{Synthetic} \\{inc. $(\mathcal{T})$}} 
				& 7.27 &  0.49& 7.64 &  0.53& \textbf{5.46} &  \textbf{0.71}& 5.91 &  0.66& 6.77 &  0.59& 7.56 &  0.47& 7.96 &  0.43\\ 
				\makecell[l]{Geometric\\ shapes $(\mathcal{P})$} 
				& 0.54 &  0.63& 0.72 &  0.84& \textbf{0.51} &  \textbf{0.79}& 0.60 &  0.77& 0.62 &  0.73& 0.77 &  0.60& 0.78 &  0.57\\ 
				\hline
				\textbf{Average} 
				& 3.90 &  0.56& 4.18 &  0.68& \textbf{2.99} &  \textbf{0.75}& 3.26 &  0.71& 3.69 &  0.66& 4.16 &  0.53& 4.37 &  0.50\\ 
				\hline
			\end{tabulary}
			%\vspace{-1cm}
		\end{center}
	\end{table}


	\bibliographystyle{splncs04}
	\bibliography{refs}
	
\end{document}